\documentclass[sigconf]{acmart}

\copyrightyear{2023}
\acmYear{2023}
\setcopyright{acmlicensed}\acmConference[MM '23]{Proceedings of the 31st
ACM International Conference on Multimedia}{October 29-November 3,
2023}{Ottawa, ON, Canada}
\acmBooktitle{Proceedings of the 31st ACM International Conference on
Multimedia (MM '23), October 29-November 3, 2023, Ottawa, ON, Canada}
\acmPrice{15.00}
\acmDOI{10.1145/3581783.3612212}
\acmISBN{979-8-4007-0108-5/23/10}

\usepackage{multirow}
\usepackage{color}
\usepackage{algpseudocode}
\usepackage{algorithm}
\usepackage[bb=boondox]{mathalfa}

\usepackage{soul} 
\usepackage{color, xcolor} 

\AtBeginDocument{%
  \providecommand\BibTeX{{%
    \normalfont B\kern-0.5em{\scshape i\kern-0.25em b}\kern-0.8em\TeX}}}

\begin{document}

\title{Towards End-to-End Unsupervised Saliency Detection with Self-Supervised Top-Down Context}

\author{Yicheng Song}
\affiliation{%
  \institution{Shanghai Key Lab of Intelligent Information Processing, School of Computer Science, Fudan University}
  \city{Shanghai}
  \country{China}}
\email{songyc22@m.fudan.edu.cn}

\author{Shuyong Gao\small*}
\affiliation{%
  \institution{Shanghai Key Lab of Intelligent Information Processing, School of Computer Science, Fudan University}
  \city{Shanghai}
  \country{China;}
  \institution{Keenon Robotics Co., Ltd.}
  \city{Shanghai}
  \country{China}
}
\email{sy_gao@fudan.edu.cn}

\author{Haozhe Xing}
\affiliation{%
	\institution{Academy for Engineering \& Technology, Fudan University}
	\city{Shanghai}
	\country{China}}
\email{hzxing21@m.fudan.edu.cn}

\author{Yiting Cheng}
\affiliation{%
  \institution{Shanghai Key Lab of Intelligent Information Processing, School of Computer Science, Fudan University}
  \city{Shanghai}
  \country{China}
}
\email{18210240023@fudan.edu.cn}

\author{Yan Wang}
\affiliation{%
	\institution{Academy for Engineering \& Technology, Fudan University}
	\city{Shanghai}
	\country{China}}
\email{yanwang19@fudan.edu.cn}

\author{Wenqiang Zhang}
\affiliation{%
	\institution{Shanghai Key Lab of Intelligent  Information Processing, School of Computer Science, Fudan University}
	\city{Shanghai}
	\country{China;}
 	\institution{Academy for Engineering \& Technology, Fudan University}
	\city{Shanghai}
	\country{China}
}
\email{wqzhang@fudan.edu.cn}
\authornote{Corresponding authors} 

\renewcommand{\shortauthors}{Yicheng Song, et al.}

\begin{abstract}

Unsupervised salient object detection aims to detect salient objects without using supervision signals eliminating the tedious task of manually labeling salient objects. 
To improve training efficiency, end-to-end methods for USOD have been proposed as a promising alternative. However, current solutions rely heavily on noisy handcraft labels and fail to mine rich semantic information from deep features. In this paper, we propose a self-supervised end-to-end salient object detection framework via top-down context. 
Specifically, motivated by contrastive learning, we exploit the self-localization from the deepest feature to construct the location maps which are then leveraged to learn the most instructive segmentation guidance. Further considering the lack of detailed information in deepest features, we exploit the detail-boosting refiner module to enrich the location labels with details.
Moreover, we observe that due to lack of supervision, current unsupervised saliency models tend to detect non-salient objects that are salient in some other samples of corresponding scenarios.
To address this widespread issue, we design a novel Unsupervised Non-Salient Suppression (UNSS) method developing the ability to ignore non-salient objects.
Extensive experiments on benchmark datasets demonstrate that our method achieves leading performance among the recent end-to-end methods and most of the multi-stage solutions. The code is available. 
\end{abstract}





\maketitle

\section{Introduction}

\begin{figure*}
  \includegraphics[width=0.9\textwidth]{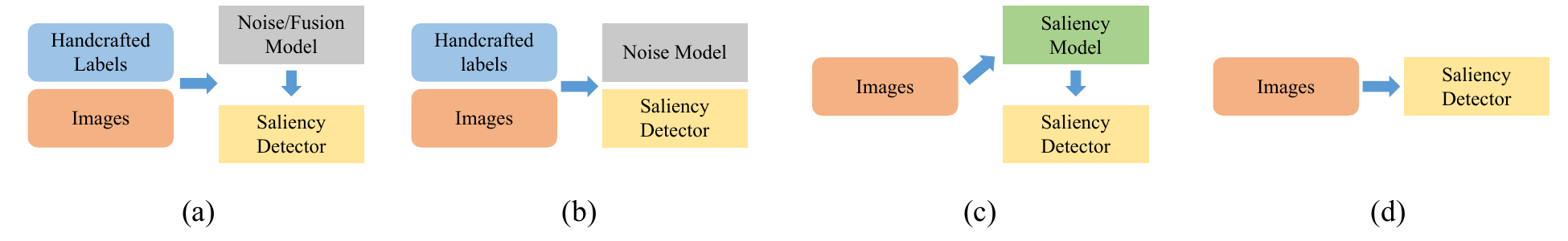}
  \caption{Pipelines of current USOD methods.  Existing learning procedures and resources are denoted by different colors. Blue: traditionally handcrafted labels; Orange: RGB images. Grey: training models for handling handcrafted labels; Green: training a model for labeling; Yellow: training a saliency detector.}
  \Description{}
  \label{fig:stage}
\end{figure*}

Unsupervised Salient Object Detection (USOD) focuses on localizing and segmenting the salient objects beyond the limits of any form of annotations. Compared to supervised methods, USOD methods are capable of applying to any dataset at a very low cost. Also, USOD methods get rid of the interactive cost of the weakly-supervised annotation, e.g., image classification\cite{piao2021mfnet} and few-pixels annotations\cite{zhang2020weakly, PSOD_aaai2022}. This enables the unsupervised approach to be easily applied to more practical scenarios, such as medical or industrial data, where there is a large amount of data and it can be difficult to label them. Meanwhile, USOD methods can assist other related tasks such as object detection \cite{diao2016efficient, patil2018msednet}, object recognition \cite{flores2019saliency}, and segmentation task \cite{flores2019saliency}. However, cumbersome steps are necessary for the current USOD methods to either learn from handcraft labels or set generating high-quality labels for the training dataset as an essential step.

Relying on the traditional salient object detection methods, a considerable part of current USOD methods\cite{zhang2017supervision, zhang2018deep, nguyen2019deepusps, zhang2020learning} train according to the non-deep learning saliency cues. Early methods learn from the one or multi-source handcraft salient maps in multiple stages. DeepUSPS\cite{nguyen2019deepusps} make refinement on distinct handcraft labels in the first stage and learn a deep neural network during the second stage. Recently, Zhou et al. \cite{zhou2022activation, zhou2023texture} develop multi-stage methods through large-scale unsupervised pre-trained networks. Based on first-stage learning for constructing reliable pseudo labels, they train the second-stage model as the final segmenter after intermediate label processing. However, multiple models are trained in multi-stage frameworks for different purposes and some methods slow down the end-to-end training, such as the widely used DenseCRF\cite{krahenbuhl2011efficient}, resulting in complicated training procedures and inefficient pipelines. 

To simplify the training program and accelerate the models' iteration, some researchers embarked on the single-stage models\cite{zhang2020learning, wang2022multi, lin2022causal}. Since current USOD methods utilize certain fixed handcraft and preprocess methods, the process of obtaining labels is usually not considered a separate stage. Zhang et al. \cite{zhang2020learning} propose to detect pure salient objects from noisy saliency maps by alternating back-propagation. Some researchers realize the limitation of interference caused by labels' noise and focus on the denoising methods. Wang et al. \cite{wang2022multi} and Lin et al. \cite{lin2022causal} achieved high-quality saliency detection with the end-to-end model but still bound by noisy labels. More or less certain signals can be extracted from annotations, yet noise is difficult to be eliminated in unsupervised methods, especially for the single-stage methods. Moreover, noisy handcraft labels are born with uncertainty and inevitably trigger a chain deviation in the subsequent processing. Although handcraft methods have been validated on multiple training datasets, these pseudo-labeling methods exhibit large performance differences across different training datasets. It is contrary to the idea that unsupervised methods can be adapted to various datasets if we should carefully pick the corresponding handcraft methods when selecting datasets.

Moreover, we find the current unsupervised methods \cite{zhou2022activation, zhou2023texture, zhang2020learning, wang2022multi, lin2022causal} pay more attention to discover the salient objects in the images but neglect to deal with the non-salient objects. The existing weakly-supervised method \cite{PSOD_aaai2022} deals with the non-salient issue by manual annotations. However, it is much more challenging to solve this problem perfectly without any supervision. The model tends to degenerate into a trivial segmenter in unsupervised situations especially when only the self-supervised strategy is employed. Since humans would typically shift their attention over different objects\cite{tian2022bi}, we argue that objects of different sizes weigh differently for humans' attention. Based on the above observation, we develop the Unsupervised Non-Salient Suppression (UNSS) for filtering the non-salient objects in the pseudo labels. 

In this paper, we design a novel end-to-end self-supervised saliency detection framework that overcomes the drawbacks of handcrafted saliency labels and multi-stage procedures to train the model with great efficiency and achieve advanced performance. Our pipeline, shown in Fig. \ref{fig:stage} (d) greatly simplifies the training procedure compared to previous approaches. 
First, we adopt a self-localizer for the given samples, from which our model learns pure salient and non-salient seed regions with extremely low noise. 
Then, the detail-boosting refiner is employed to mine rich semantics including boundaries and textures. 
The self-localizer and the detecter are trained simultaneously as a complement to each other. 
To explore more accurate pseudos, we propose an effective strategy for non-salient suppression in the unsupervised training situation. Finally, our method produces a high-performance salient object detector in straightforward end-to-end training.

The main contributions of this paper can be summarized as follows:
\begin{itemize}

\item We propose a novel end-to-end self-supervised salient object detection framework that employs an unsupervised single-stage manner to efficiently train the model and produce high-quality saliency maps.

\item We design a top-down context guidance strategy that extracts properly detailed signals for global and local segmentation learning. Moreover, we propose the first unsupervised non-salient suppression method and improve the USOD accuracy. 

\item Extensive experiments on five widely used benchmarks show our method achieves state-of-the-art end-to-end USOD performance.

\end{itemize}

\section{Related Work}

\begin{figure*}
  \includegraphics[width=0.9\textwidth]{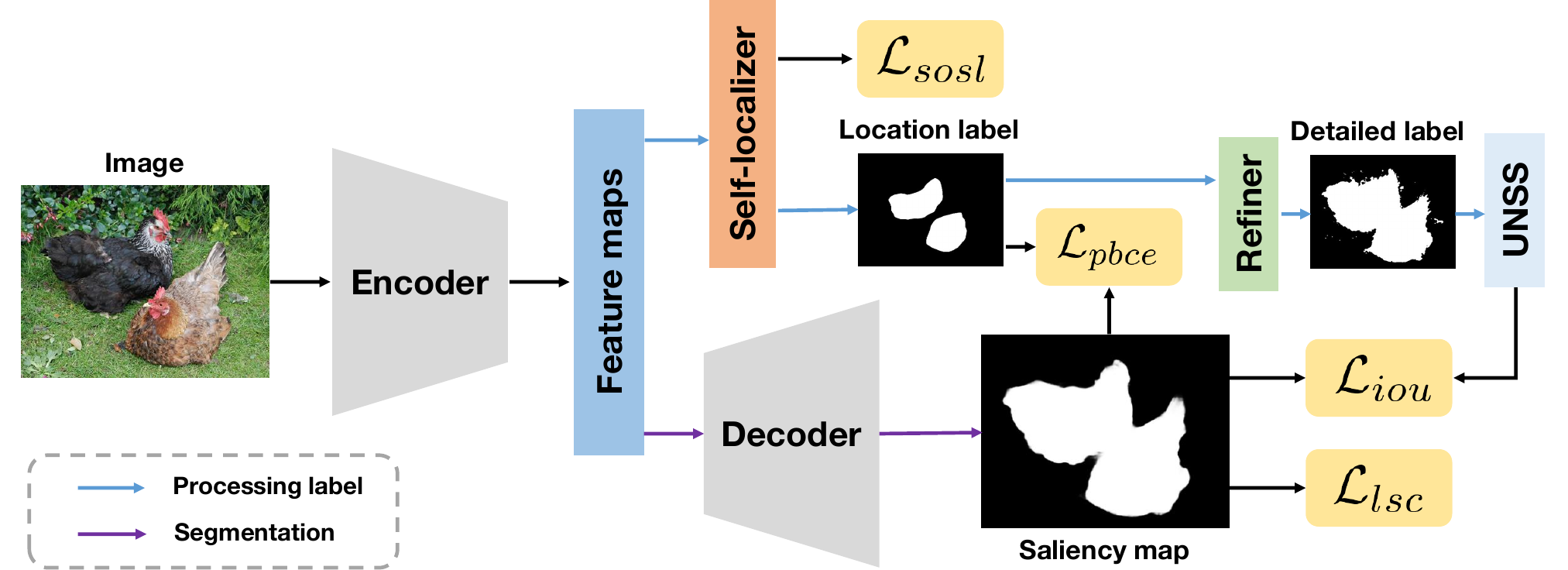}
  \caption{Proposed end-to-end USOD model architecture. The encoder extracts feature maps and the decoder produces saliency maps. Our model is supervised by two-level pseudos including location labels and detailed labels through top-down context. The proposed UNSS is employed for enhancing salient semantics.}
  \Description{}
  \label{fig:arch}
\end{figure*}

\subsection{Supervised Salient object Detection}
\textbf{Fully-supervised Method.} Significant progress has been achieved in the field of Salient Object Detection (SOD) due to the development of deep learning techniques and large-scale benchmark datasets \cite{shi2015hierarchical,li2014secrets,li2015visual,russakovsky2015imagenet,yang2013saliency}. Early SOD methods \cite{wang2015deep,lee2016deep} used multilayer perception to identify salient regions in images, which are inefficient in capturing the spatial information of images and require a lot of time to process. Then, full convolution network-based methods, such as \cite{liu2019simple, qin2019basnet, pang2020multi} became dominant and greatly improved the performance of the model in terms of accuracy and speed. While fully supervised SOD methods, however, are limited by the need for large-scale pixel-level labeled training data, which can be impractical and expensive to acquire.

\textbf{Semi-/Weakly-supervised Method.}
To alleviate the burden of pixel-level manual annotation, researchers have investigated the use of weak supervision, including category label \cite{wang2017learning}, scribbles annotation \cite{zhang2020weakly}, point annotation \cite{PSOD_aaai2022}, image caption \cite{zeng2019multi}, etc. For example, in order to capture potentially salient regions, Wang et al. \cite{wang2017learning} design a foreground inference network (FIN) which is trained by leveraging the category labels and jointly optimized with a fully convolutional network. \cite{zhang2020weakly} use scribbles annotation to give local supervision signal while utilizing auxiliary edge to complement detail information. \cite{PSOD_aaai2022} use point annotation to provide location information of salient objects to obtain pseudo labels to provide the first round of supervision, and then again use point annotation to suppress non-salient objects to obtain optimized pseudo labels for the second round of training. The above approaches employ a relatively time-efficient manual label training model to obtain a trade-off between cost and performance and utilize the two-stage paradigm for the training approach in order to improve the quality of the saliency map.

\subsection{Unsupervised Salient Object Detection}

\textbf{End-to-End Methods.} Recent USOD methods make great progress in detection performance against traditional handcraft methods\cite{jiang2013saliency, li2013saliency, yan2013hierarchical, zhu2014saliency}. Since the lack of saliency information, most current end-to-end methods\cite{zhang2020learning, wang2022multi, lin2022causal} extracted useful cues from the handcrafted labels, the typical pipeline is shown in Fig. \ref{fig:stage} (b). SBF \cite{zhang2017supervision} first used deep modeling method to learn from multiple handcraft salient maps. EDNS\cite{zhang2020learning} set salient detection as a subtask to product saliency map with noise. UMNet\cite{wang2022multi} worked on multi-source handcrafted labels and modeled the uncertainty noise from labels. DCFD\cite{lin2022causal} find causal debiasing in handcraft methods in order to learn better from single-source handcraft labels than multi-source. However, these works are limited by noisy handcraft labels and ignore the rich contrastive semantic information between images. 

\textbf{Multi-stage Methods.} Many methods\cite{zhang2017supervision, zhang2018deep, nguyen2019deepusps, shin2022unsupervised, zhou2022activation, zhou2023texture} are proposed to tackle the low accuracy in one-stage methods. As shown in Fig \ref{fig:stage} (a), early multi-stage unsupervised methods pursue a reliable preprocess before detector training. DeepUSPS \cite{nguyen2019deepusps} used multi-stage to obtain saliency maps from all kinds of handcraft labels. Selfmask \cite{shin2022unsupervised} succeeded to get rid of the noisy labels but extract the activation maps from the image samples themselves. With the development of self-supervision, recent works discover trustworthy preprocess methods as shown in Fig. \ref{fig:stage} (c). A2S \cite{zhou2022activation} and TSD\cite{zhou2023texture} started to use the self-supervised pre-trained model to make high-quality pseudo labels for the training set which is strong supervision to produce better performance in a second-stage training. However, these methods need heavy work in processing labels for the training set and rely on mid-process methods like DenseCRF \cite{krahenbuhl2011efficient} which is time-consuming in the whole pipeline. 

\subsection{Contrastive learning.} Contrastive learning is aim to extract robust semantic representation without any annotations by pulling close the positive samples and pushing apart the negative samples \cite{chen2020simple, chopra2005learning, hadsell2006dimensionality, he2020momentum, schroff2015facenet}. Based on object class labels, the positive pair is created with the samples from the same class, while the samples with different class labels form the negative pair \cite{chopra2005learning, hadsell2006dimensionality, khosla2020supervised, schroff2015facenet}. Unsupervised contrastive learning can be divided into instance-wise contrastive learning \cite{chen2020simple, he2020momentum, wu2018unsupervised, ye2019unsupervised} and clustering-based contrastive learning \cite{liprototypical, xie2016unsupervised, yang2016joint}. Instance-wise contrastive learning \cite{chen2020simple,he2020momentum} constructs the positive pairs by collecting samples from the same instances, and negative pairs are created from distinctive instances. For clustering-based contrastive learning, training samples are produced from clustering algorithm\cite{ge2020self}. As a result, a growing number of downstream unsupervised tasks\cite{oord2018representation, chen2020simple, bielskimove, wang2021exploring} benefit from powerful self-supervised visual representation.

\section{Methodology}

\subsection{Architecture}

To improve performance and efficiency, we propose an end-to-end self-supervised salient object detection model as shown in Fig. \ref{fig:arch}. Our model balances the top-level context in the global view and the bottom-level context in local details. For given a batch of images ${\{\mathbf{X}_i\}}_{i=1}^n$, the encoder $\operatorname{E}(\cdot)$ extract the multiple feature maps denoted as ${\{\mathbf{F}_i\}}_{i=1}^m$. To capture the most salient semantics, the class-agnostic activation map \cite{Xie_2022_CVPR} is used to produce self-localization for generating the location labels denoted as $\mathbf{G}$ without any supervision. Location labels $\mathbf{G}$ mark a distinctive region of the foreground against the background with deep semantic features. 
The pixel-wise refiner \cite{ru2022learning} produces detailed labels $\mathbf{G}_r$ to enhance semantic features using location labels $\mathbf{G}$ and the images. In order to tackle the non-salient objects detected by localizer or refiner, we propose Unsupervised Non-Salient Suppression (UNSS) to rectify the fading saliency information. A multi-task loss function is employed for different constraint purposes. To generate valid self-visual representation, $\operatorname{E}(\cdot)$ is initialized with the unsupervised pretraining moco\cite{he2020momentum}, which is trained in ImageNet-1k\cite{russakovsky2015imagenet} without any manual annotations. 

\subsection{Salient Object Self-Localizer}

High activation features generated from large-scale unsupervised pre-trained networks are studied by Zhou et al. \cite{zhou2022activation, zhou2023texture}, which motivates us to build contrastive learning between those distinctive features. In order to extract accurate saliency knowledge and get rid of the limitation of handcrafted labels, we adopt Class-agnostic Activation Map (C2AM)\cite{Xie_2022_CVPR} as the salient object self-localizer to make the location labels. Unlike widely-used Class Activation Map (CAM) \cite{zhou2016learning}, the class-agnostic activation map fuses the objects' classification and focuses on foreground-background contrast. To generate deep semantic features, ResNet-50\cite{he2016deep} is employed as our backbone encoder. So we get five groups of feature maps from the encoder $\operatorname{E}(\cdot)$. Given a batch of images $\mathbf{X}$, we produce the initial class-agnostic activation map $\mathbf{G}_o$ by:
\begin{equation}
\{\mathbf{F}_1, \mathbf{F}_2, \mathbf{F}_3, \mathbf{F}_4, \mathbf{F}_5\}=\operatorname{E}(\mathbf{X})
\end{equation}
\begin{equation}
 \mathbf{G}_o=\operatorname{Sigmoid}(\operatorname{BN}(\sum_{i=1}^{d}\mathbf{W}_{i}(\mathbf{F}_4 \oplus \mathbf{F}_5)))
\end{equation}
where $\oplus$ denotes concatenation, ${\{\mathbf{W}_i\}}_{i=1}^d$ denote the $d$ channels' weights in convolution layer, $\operatorname{BN}(\cdot)$ denotes the batch normalization layer, $\operatorname{Sigmoid}(\cdot)$ is the sigmoid activation function. To generate robust foreground labels, we produce the self-localization map $\mathbf{G}$ by class-agnostic activation maps $\{\mathbf{G}_o, \mathbf{G}_o^{\prime}, \mathbf{G}_o^{\prime \prime}\}$ from input images in three different sizes:
\begin{equation}
\mathbf{G}=\operatorname{maxpool}(\mathbf{G}_o, \mathbf{G}_o^{\prime}, \mathbf{G}_o^{\prime \prime})
\end{equation}
where $\operatorname{maxpool}(\cdot)$ denotes the max pooling layer. 

\textbf{Foreground-background similarity.} The idea of contrastive learning is to construct positive and negative sample pairs and keep them at suitable distances from each other. We use a batch of activation maps and measure the similarity between foreground samples and background samples. For a batch of $n$ class-agnostic activation maps $\{\mathbf{G}_i\}_{i=1}^n$ and feature maps $\{\mathbf{F}_i\}_{i=1}^5$, we construct opposite samples for each channel:
\begin{equation}
\mathbf{s}_i^f=\mathbf{G}_i \otimes \mathbf{F}_i^{\top}, \quad \mathbf{s}_i^b=\left(\mathbf{1}-\mathbf{G}_i\right) \otimes \mathbf{F}_i^{\top}
\end{equation}
where in our case, the positive samples $\mathbf{s}^f$ denote the foreground samples, and $\mathbf{s}^b$ denote background samples. The positive samples are assumed as foreground at first and deduced after the warming-up epoch. We use three groups of cross-image similarity contrast to describe the distance between samples. For different channel $i$ and $j$, we construct inter-foreground similarity $\mathbf{S}_{i,j}^f$, inter-background similarity $\mathbf{S}_{i,j}^b$, and foreground-background similarity $\mathbf{S}_{i,j}^{fb}$:
\begin{equation}
\mathbf{S}_{i,j}^f=\operatorname{sim}(\mathbf{s}_i^f, \mathbf{s}_j^f), \quad \mathbf{S}_{i,j}^b=\operatorname{sim}(\mathbf{s}_i^b, \mathbf{s}_j^b), \quad \mathbf{S}_{i,j}^{fb}=\operatorname{sim}(\mathbf{s}_i^f, \mathbf{s}_j^b)
\end{equation}
where $\operatorname{sim}(\cdot)$ denotes the cosine similarity. 

\textbf{Contrastive learning loss.} To pull close the foreground samples and pull apart the background samples, we minimize the contrastive learning loss between samples. Given the similarity measurements for a batch of feature maps, we construct foreground-background similarity set $\mathbf{S}^{fg}=\{\mathbf{S}_{1,1}^{fg}, \cdots, \mathbf{S}_{i,j}^{fg}, \cdots, \mathbf{S}_{n,n}^{fg}\}$ and calculate the loss of the negative pairs by:
\begin{equation}
\mathcal{L}_{N E G}=-\frac{1}{\lVert \mathbf{S}^{fg} \rVert} \sum_{\mathbf{S}_k \in \mathbf{S}^{fg}} \log \left(1-\mathbf{S}_{k}\right)
\end{equation}
We pull closer the positive samples both in foreground samples and background samples. For the cross-image foreground similarity set $\mathbf{S}^f=\{\mathbf{S}_{1,2}^f, \cdots, \mathbf{S}_{i,j}^f, \cdots \}(i\neq j)$ and the cross-image background similarity set $\mathbf{S}^g=\{\mathbf{S}_{1,2}^g, \cdots,  \mathbf{S}_{i,j}^g, \cdots\}(i\neq j)$. Generally, we use positive contrastive learning loss $\mathcal{L}_{P O S}(\cdot)$ to product positive loss for $\mathbf{S}\in \{\mathbf{S}^f, \mathbf{S}^g\}$ by:
\begin{equation}
\mathcal{L}_{POS}(\mathbf{S})=-\frac{1}{\lVert \mathbf{S} \rVert} \sum_{\mathbf{S}_k \in \mathbf{S}} \mathbf{H}(\mathbf{S}_{k}) \cdot \operatorname{log}(\mathbf{S}_{k})
\end{equation}
where we use $\operatorname{H}(\cdot )$ to control the weight of the positive loss using the reliability ranking for each score in the similarity set. $\operatorname{H}(\cdot )$ is defined as:
\begin{equation}
\operatorname{H}(\mathbf{S}_{k}) = \exp \left(-\alpha \cdot \operatorname{rank}(\mathbf{S}_{k})\right)
\end{equation}
In the equation, $\alpha$ controls the smoothness of the ranking, $rank(\cdot )$ sort the similarities and return the order of each score. $\operatorname{R}(\mathbf{S}_{k})$ helps similar samples to make greater contributions to supervise the parameters. Finally, the total loss $\mathcal{L}_{sosl}$ is used to gather the similar features and separate the dissimilar features:
\begin{equation}
\mathcal{L}_{sosl} = \mathcal{L}_{P O S}(\mathbf{S}^f) + \mathcal{L}_{P O S}(\mathbf{S}^g) + \mathcal{L}_{N E G}
\end{equation}

\subsection{Salient Detail Refiner}
End-to-end methods are not able to directly obtain detailed information from pictures in training procedures compared with multi-stage methods with intermediate processors like DenseCRF \cite{krahenbuhl2011efficient}. Inspired by \cite{ru2022learning}, we use real-time pixel refiner to find local detail around the location labels $G$. Since the local position and pixel-wise features are vital information to optimize feature details, we define the feature distance $d_f^{i, j}$ and position distance $d_p^{i, j}$ between pixels $\mathbf{G}_i$ and $\mathbf{G}_j$ coordinated in the spatial location:
\begin{equation}
d_f^{i, j} = -{(\frac{\lVert \mathbf{G}^{i} - \mathbf{G}^{j}\rVert}{\gamma_1 \sigma_f})}^2, 
d_p^{i, j} = -{(\frac{\lVert \vec i - \vec j\rVert}{\gamma_2 \sigma_p})}^2
\end{equation}
where $\sigma_f$ and $\sigma_p$ mean the standard deviation of feature value and position difference. $\gamma_1$ and $\gamma_2$ control the smoothness. For single-position refinement, we define refiner $\operatorname{R}(\cdot)$ as:
\begin{equation}
\operatorname{R}(\mathbf{G}^i) = \sum_{j \in \mathcal{N}(i)} (
        \frac{exp(d_f^{i, j})}{\sum_{k \in \mathcal{N}(i)} exp(d_f^{i, k})} +
        \gamma_3 \frac{exp(d_p^{i, j})}{\sum_{k \in \mathcal{N}(i)} exp(d_p^{i, k})}
    )
\end{equation}
Here, $\mathcal{N}(\cdot)$ denotes the 8-way neighbor pixel set. The refiner merges the result and remains the local consistency from the origin pseudo label in several iterations using the following function:
\begin{equation}
    {(\mathbf{G}_{r}^{i})}_{t} = \operatorname{R}({(\mathbf{G}_{r}^{i})}_{t-1}) \odot {(\mathbf{G}_{r}^{i})}_{t-1}
\end{equation}
where $\odot$ denotes the Hadamard product for matrixes. 

\subsection{Unsupervised Non-Salient Suppression}

During the processing of pseudos, some non-salient objects emerge since the sparsity of the labels. The non-salient objects are highlighted either during self-localization since the relatively distinguished features against the background or during the refinement because of the local similarity in the images. 

\begin{figure}[!t]
\centering
\includegraphics[width=0.48 \textwidth]{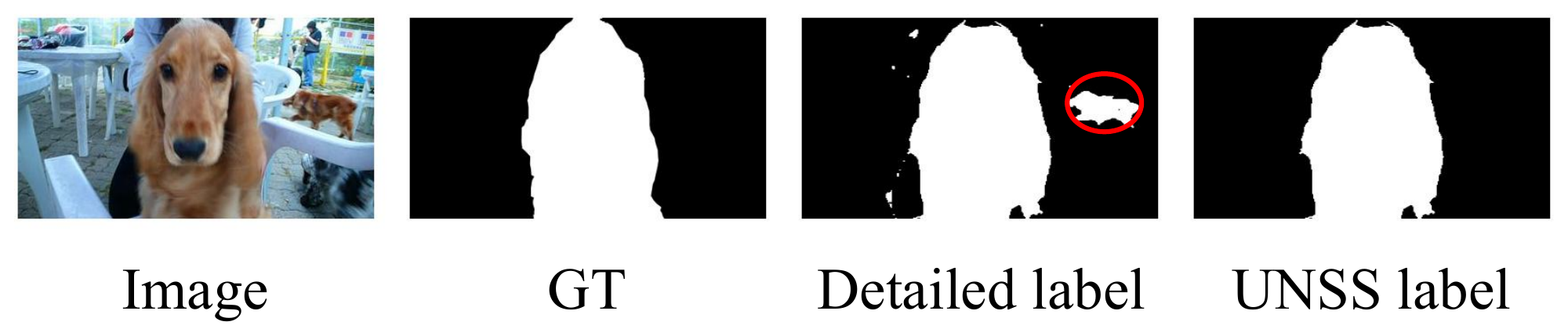}
\caption{Illustration of the UNSS method. The small object in the red circle is suppressed, and the remaining object is much closer to GT. }
\label{fig:unss}
\end{figure}

\begin{algorithm}
\caption{Unsupervised Non-Salient suppression}\label{alg:unss}
\begin{flushleft}
\end{flushleft}
\begin{algorithmic}[1]
\State \textbf{procedure} UNSS($\mathbf{G}_r$, $\theta_r$)
\State \quad sort objects in $\mathbf{G}_r$ by decreasing area
\State \quad $\mathbf{G_{nss}} \leftarrow \emptyset$
\State \quad $n \leftarrow \lVert \mathbf{G}_r \rVert$
\State \quad \textbf{for} $\mathbf{g}_i \in \mathbf{G_{r}}$ \textbf{do}
\State \quad \quad \textbf{if} $i = n$ or $\operatorname{area}(\mathbf{g}_i) \le \operatorname{area}(\mathbf{g_{i+1}}) \times \theta_r$ \textbf{then}
\State \quad \quad \quad $\mathbf{G_{nss}} \leftarrow \mathbf{G_{nss}} \cup \mathbf{g}_i$
\State \quad \quad \textbf{else}
\State \quad \quad \quad \textbf{break}
\State \quad $\mathbf{G_{r}} \leftarrow \mathbf{G_{nss}}$
\State \quad \textbf{return} $\mathbf{G_{r}}$
\end{algorithmic}
\end{algorithm}

To avoid the dying away of salient signals, we propose the Unsupervised Non-Salient Suppression (UNSS) method to enhance the supervision. As shown in Fig. \ref{fig:unss}, the detailed labels highlight the non-salient object pseudo either from mistaken localization or from local similarity. However, non-salient objects are always incomplete or in very small size compare to the salient object while they share similar semantics. As a result, human tends to ignore objects that are proportionally smaller than other objects. Our UNSS method is based on the size proportion between detected objects in the labels. In Algorithm \ref{alg:unss}, we first sort all objects by decreasing sizes and use threshold $\theta_r$ to compare the size proportion of two objects. Those objects are suppressed if they are in quite small size compared to others. 

As shown in Fig. \ref{fig:unss}, the objects in a small size are ignored in the comparison of the objects in a large size. Meanwhile, small background regions that hold similar semantics to the salient objects can be also ignored if boundaries are detected by the refiner, which further improves accuracy.

\subsection{End-to-End Training}

We define a multi-task learning loss function for several purposes. Inspired by the loss design in PSOD\cite{PSOD_aaai2022}, we employed partial cross-entropy loss\cite{tang2018normalized} and local structure-consistent loss\cite{yu2021structure} to learn accurate information from pseudo labels. Also, the IOU loss is used for minimizing the error in predictions. 

For the decoder training in the pipeline, the partial binary cross-entropy loss is used to learn from certain foreground and background areas. For the location label $\mathbf{G}$, the certain label is created by:
\begin{equation}
\mathbf{G}^{i}=\left\{\begin{array}{lr}
1, & \text { if } \mathbf{G}^{i}\geq \theta_f \\
0, & \text { if } \mathbf{G}^{i}\leq \theta_g
\end{array}\right.
\end{equation}
where $\theta_f$ is the threshold to pick out the foreground area and $\theta_g$ is the low threshold to pick out the background area. Focusing on the area in $\mathbf{G}$, partial binary cross-entropy is obtained by:
\begin{equation}
\mathcal{L}_{pbce}=-\sum_{i \in \mathbf{M}}\left[\mathbf{G}^i \log \left(\mathbf{M}^i\right)+\left(1-\mathbf{G}^i\right) \log \left(1-\mathbf{M}^i\right)\right]
\end{equation}
where $\mathbf{G}$ is the location label from the self-localizer and $M$ is the saliency map. To keep the local consistency in the saliency map, local structure-consistent is used:
\begin{equation}
\mathcal{L}_{lsc}=\sum_{i}\sum_{j\in K_i}|\mathbf{M}^i-\mathbf{M}^j|\operatorname{f}(i, j)
\end{equation}
where $K_i$ is the areas covered by $k\times k$ kernel around pixel $i$. $|\cdot|$ denotes the L1 distance between the salient value on two pixels. And $\operatorname{f}(i, j)$ is the Gaussian kernel bandwidth filter following the setting in PSOD\cite{PSOD_aaai2022}.

In order to reduce the error in predictions, the IOU loss is adopted for supervision. Using the detailed label $G_r$ from the refiner, the IOU loss is calculated by:

\begin{equation}
\mathcal{L}_{iou}=1-\frac{\lVert \mathbf{M}\cap \mathbf{G}_r\rVert}{\lVert \mathbf{M}\cup \mathbf{G}_r\rVert}
\end{equation}
here $\lVert \cdot \rVert$ denotes the norm for the pixels set. So the loss we used to supervise the end-to-end model is defined as:

\begin{equation}
\mathcal{L}=\alpha \mathcal{L}_{sosl} + \beta_1 (\mathcal{L}_{pbce} + \mathcal{L}_{lsc}) + \beta_2 \mathcal{L}_{iou}
\end{equation}
where $\alpha$, $\beta_1$, and $\beta_2$ are weights to balance the different losses.

\begin{table*}[t]
  \centering
\renewcommand\tabcolsep{3.5pt}
\vspace{-0.1in}
  \caption{Experiments for SOD benchmark datasets measured in ave-$F_\beta$, $E_{\xi}$, and MAE metrics. $\uparrow$ and $\downarrow$ indicate that the larger and smaller scores are better, respectively. $\dagger$ denotes the models trained in the MSRA-B dataset \cite{liu2010learning}. \textbf{Bold} numbers indicate the best performance in each group.}
  \label{tab:benchmark}
    \begin{tabular}{c|c|ccc|ccc|ccc|ccc|ccc}
    \toprule
          \multirow{2}[1]{*}{Methods} & \multirow{2}[1]{*}{Year} & \multicolumn{3}{c|}{ECSSD} & \multicolumn{3}{c|}{DUT-O} & \multicolumn{3}{c|}{PASCAL-S} & \multicolumn{3}{c|}{DUTS-TE} & \multicolumn{3}{c}{HKU-IS} \\
          &       & $F_\beta \uparrow$ & $E_{\xi} \uparrow$ & $\mathcal{M} \downarrow$  & $F_\beta \uparrow$ & $E_{\xi} \uparrow$ & $\mathcal{M} \downarrow$  & $F_\beta \uparrow$ & $E_{\xi} \uparrow$ & $\mathcal{M} \downarrow$  & $F_\beta \uparrow$ & $E_{\xi} \uparrow$ & $\mathcal{M} \downarrow$  & $F_\beta \uparrow$ & $E_{\xi} \uparrow$ & $\mathcal{M} \downarrow$\\
    \midrule
    \multicolumn{17}{c}{Fully/Weakly-supervised models} \\
    \midrule
    WSSA\cite{zhang2020weakly}  & 2020  & 0.870  & 0.917 & 0.059 & 0.703 & 0.845 & 0.068 & 0.785 & 0.855 & 0.096 & 0.742 & 0.869 & 0.062 & 0.860  & 0.932 & 0.047 \\
    MFNet\cite{piao2021mfnet} & 2021  & 0.844 & 0.889 & 0.084 & 0.621 & 0.784 & 0.098 & 0.756 & 0.824 & 0.115 & 0.693 & 0.832 & 0.079 & 0.839 & 0.919 & 0.058 \\
    SCW\cite{yu2021structure}   & 2021  & 0.900   & 0.931 & 0.049 & 0.758 & 0.862 & 0.06  & 0.827 & 0.879 & 0.080  & 0.823 & 0.890  & 0.049 & 0.896 & 0.943 & 0.038 \\
    
    BASNet \cite{qin2019basnet} & 2019 &0.879 &0.921 &0.037 &0.756 &0.869 &0.056 &0.781 &0.853 &0.077 &0.791 &0.884 &0.048 &0.898 &0.947 &0.033 \\
    CPD \cite{wu2019cascaded} & 2019 &0.917 &0.949 &0.037 &0.747 &0.873 &0.056 &0.831 &0.887 &0.072 &0.805 &0.904 &0.043 &0.891 &0.952 &0.034 \\
    ITSD \cite{zhou2020interactive} & 2020 &0.905 &0.933 &0.035 &0.750 &0.863 &0.059 &0.817 &0.868 &0.066 &0.808 &0.897 &0.040 &0.899 &0.953 &0.030 \\
    MINet\cite{pang2020multi} & 2020  & \textbf{0.924} & \textbf{0.953} & \textbf{0.033} & \textbf{0.756} & \textbf{0.873} & \textbf{0.055} & \textbf{0.842} & \textbf{0.899} & \textbf{0.064} & \textbf{0.828} & \textbf{0.917} & \textbf{0.037} & \textbf{0.908} & \textbf{0.961} & \textbf{0.028} \\
    \midrule
    \multicolumn{17}{c}{Multi-Stage unsupervised models} \\
    \midrule
    SBF \cite{zhang2017supervision}$\dagger$ & 2017 & 0.812 & 0.878 & 0.087 & 0.611 & 0.771 & 0.106 & 0.711 & 0.795 & 0.131 & 0.627 & 0.785 & 0.105 & 0.805 & 0.895 & 0.074 \\
    MNL \cite{zhang2018deep}$\dagger$ & 2018 & 0.874 & 0.906 & 0.069  & 0.683 & 0.821 & 0.076 & 0.792 & 0.846 & 0.091 & - & - & - & 0.874 & 0.932 & 0.047 \\
    DeepUSPS\cite{nguyen2019deepusps}$\dagger$ & 2019  & 0.875 & 0.903 & 0.064 & 0.715 & 0.839 & 0.116 & 0.770  & 0.828 & 0.107 & 0.730  & 0.840  & 0.072 & 0.880  & 0.933 & 0.043 \\
    SelfMask\cite{shin2022unsupervised} & 2022  & 0.856 & 0.920  & 0.058 & 0.668 & 0.815 & 0.078 & 0.774 & 0.856 & 0.087 & 0.714 & 0.848 & 0.063 & 0.819 & 0.915 & 0.053 \\
    A2S\cite{zhou2022activation} & 2023 & 0.882 & 0.921 & 0.056 & 0.688 & 0.818 & 0.079 & 0.778 & 0.842 & 0.100 & 0.729 & 0.847 & 0.069 & 0.868 & 0.936 & 0.041 \\
    TSD\cite{zhou2023texture}   & 2023  & \textbf{0.916} & \textbf{0.938} & \textbf{0.044} & \textbf{0.745} & \textbf{0.863} & \textbf{0.061} & \textbf{0.830}  & \textbf{0.882} & \textbf{0.074} & \textbf{0.810}  & \textbf{0.901} & \textbf{0.047} & \textbf{0.902} & \textbf{0.947} & \textbf{0.037} \\
    \midrule
    \multicolumn{17}{c}{End-to-End unsupervised models} \\
    \midrule
    EDNS\cite{zhang2020learning}  & 2020  & 0.872 & 0.906 & 0.068 & 0.682 & 0.821 & 0.076 & 0.801 & 0.846 & 0.097 & 0.735 & 0.847 & 0.065 & 0.874 & 0.933 & 0.046 \\
    UMNet\cite{wang2022multi}$\dagger$  & 2022  & 0.879 & 0.904 & 0.064 & 0.743 & \textbf{0.860}  & \textbf{0.063} & 0.770  & 0.830  & 0.106 & 0.752 & 0.863 & 0.067 & 0.889 & 0.940  & \textbf{0.041} \\
    DCFD\cite{lin2022causal}  & 2022  & 0.888 & 0.915 & 0.059 & 0.710  & 0.837 & 0.070  & 0.795 & 0.860  & 0.090  & 0.764 & 0.855 & 0.064 & 0.889 & 0.935 & 0.042 \\
    Ours  & - & \textbf{0.903} & \textbf{0.935} & \textbf{0.050} & \textbf{0.753} & 0.852 & 0.068 & \textbf{0.827} & \textbf{0.881} & \textbf{0.076} & \textbf{0.809} & \textbf{0.891} & \textbf{0.052} & \textbf{0.891} & \textbf{0.942} & \textbf{0.041} \\
    \bottomrule
    \end{tabular}%
  \label{tab:addlabel}%
\end{table*}%

\section{Experiments}

\textbf{Implementation details.} We implemented our model with Pytorch and all experiments are done on a single TITAN Xp GPU. In our loss function, $\alpha$ is set to 1, $\beta_1$ and $\beta_2$ is set to 0 in the warming-up epoch, after that the $\alpha$ is set to 0.1, $\beta_1$ is set to 1, and $\beta_2$ is set to 0.1 in the following epochs. We set the foreground threshold $\theta_f$ to 0.6 and the background threshold $\theta_g$ to 0.1 for generating both location labels and detailed labels. For label refiner, we set smoothness $\gamma_1$ and $\gamma_2$ to both 0.4 and position weight $\gamma_3$ to 0.01. For the hyperparameter used in UNSS, we set $\theta_r$ to 2.5 for best performance. To conduct contrastive learning and avoid the degradation of the model, we use horizontal flipping and random cropping to enhance our data. The input images are resized to $352 \times 352$. The total training process is completed in 10 epochs setting the batch size to 16, we employ the SGD optimizer with a learning rate of 0.005 for the decoder, and 0.0005 for both the encoder and self-localizer. During model testing, we resized each image to $352 \times 352$ to predict the final saliency maps without any post-processing. 

\textbf{Datasets.} Following the prior work, we train our model on DUTS\cite{russakovsky2015imagenet} dataset without any annotation. All experiments are finished on five widely-used datasets for evaluation, including
ECSSD\cite{shi2015hierarchical}, PASCAL-S\cite{li2014secrets}, HKU-IS\cite{li2015visual}, DUTS-TE\cite{russakovsky2015imagenet}, DUT-O\cite{yang2013saliency}. 

\textbf{Metrics.} For producing a fair comparison, we take three criteria for evaluation, including ave-$F_\beta$, Mean Absolute Error (MAE), and E-Measure ($E_{\xi}$) \cite{fan2018enhanced}. F-measure is the evaluation of the results both on precision and recall, which is defined as: 
\begin{equation}
    F_\beta=\frac{(1+\beta^2)\times Precision \times Recall}{\beta^2\times Precision + Recall}
\end{equation}
where $\beta^2$ is set to 0.3 \cite{achanta2009frequency}. The ave-$F_\beta$ is the $F_\beta$ score by setting the threshold as two times the mean values. MAE is the absolute error between predictions and ground truth. $E_{\xi}$ measures the global statistics and pixel-matching information.

\begin{figure}
\centering
\includegraphics[width=0.44 \textwidth]{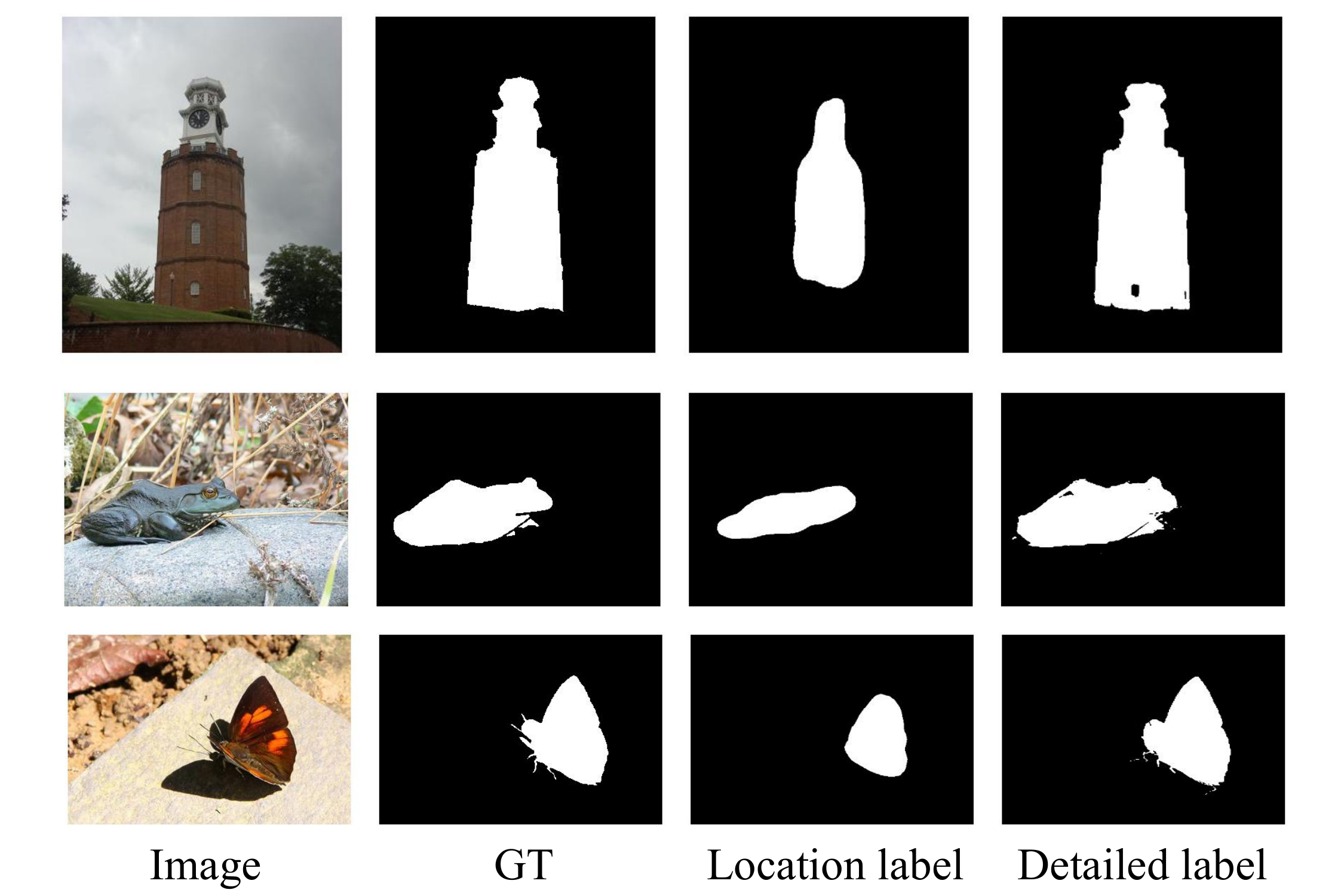}
\caption{Examples of the generated pseudo labels. Location labels generate coarse-grained saliency maps and detailed labels learn more specific semantics.}
\label{fig:refinement_view}
\end{figure}

\begin{figure*}
  \includegraphics[width=0.8\textwidth]{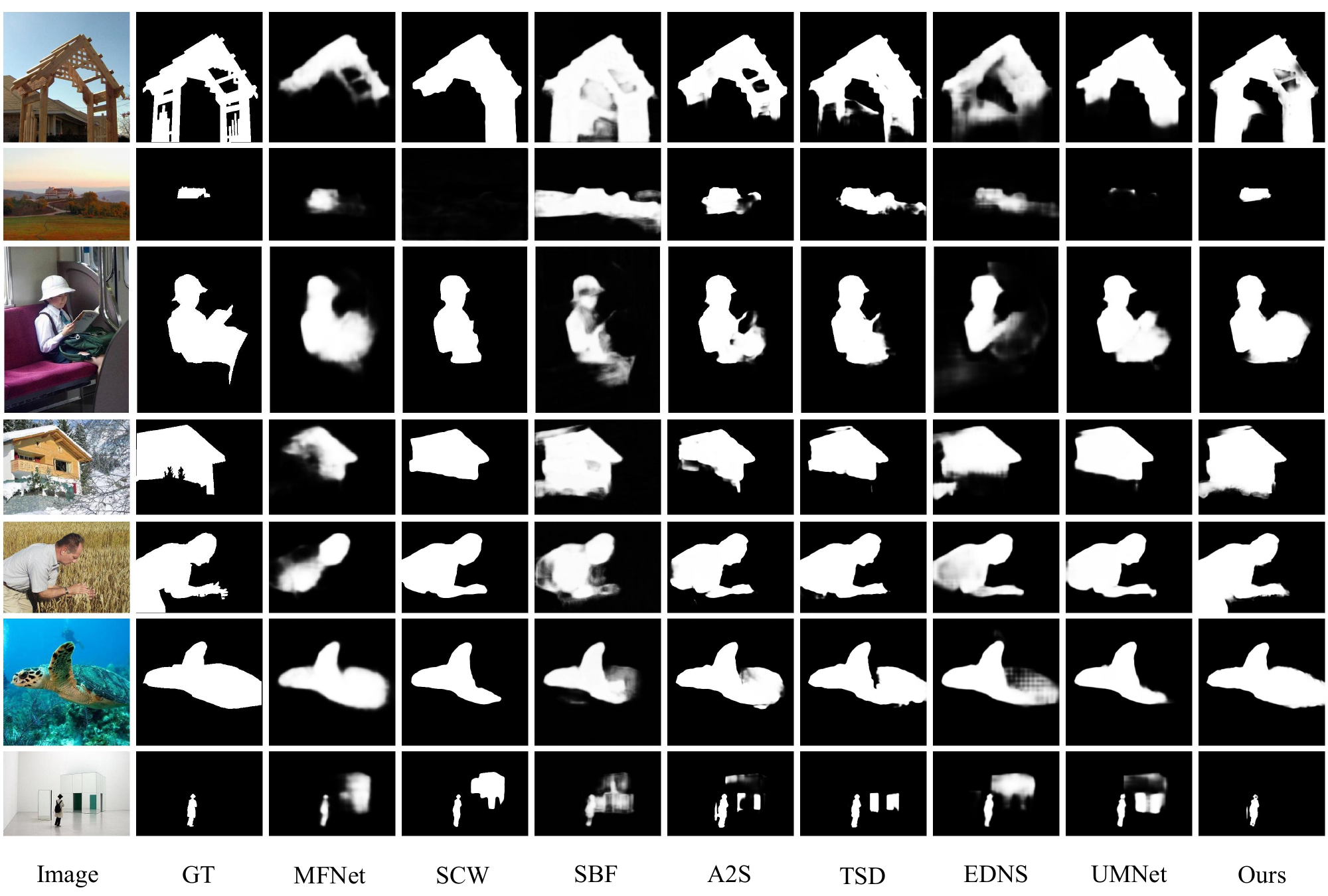}
  \caption{Visual comparison on different methods. Columns 3 and 4 are weakly supervised methods, columns 5 to 7 are multi-stage unsupervised methods, and columns 8 and 9 are single-stage unsupervised methods.}
  \Description{}
  \label{fig:visual_contrast}
\end{figure*}

\subsection{Comparison to state-of-the-art}

\begin{table}
  \caption{Label quality on DUTS-TR dataset.}
  \label{tab:refinement}
  \begin{tabular}{l|ccc}
    \toprule
    Label type & $F_\beta \uparrow$ & $E_{\xi} \uparrow$ & $\mathcal{M} \downarrow$ \\
    \midrule
    Location label & 0.827 & 0.766 & 0.130 \\
    Detailed label & 0.887 & \textbf{0.920} & 0.060 \\
    UNSS label & \textbf{0.891} & \textbf{0.920} & \textbf{0.059} \\
    \bottomrule
  \end{tabular}
\end{table}

\textbf{Quantitative evaluation.} As shown in Tab. \ref{tab:benchmark}, our model is compared with fully-supervised methods including BASNet \cite{qin2019basnet}, CPD \cite{wu2019cascaded}, ITSD \cite{zhou2020interactive} and MINet\cite{pang2020multi}, weakly-supervised methods including WSSA\cite{zhang2020weakly}, MFNet\cite{piao2021mfnet}, SCW\cite{yu2021structure}, multi-stage unsupervised methods including SBF \cite{zhang2017supervision}, MNL \cite{zhang2018deep}, DeepUSPS\cite{nguyen2019deepusps}, SelfMask\cite{shin2022unsupervised}, A2S\cite{zhou2022activation}, TSD\cite{zhou2023texture} and single-stage unsupervised methods including EDNS\cite{zhang2020learning}, UMNet\cite{wang2022multi}, DCFD\cite{lin2022causal}. We separate our benchmark for multi-stage and single-stage unsupervised methods because of the huge overhead gap between the two kinds of methods. In our benchmark, our model outperforms the state-of-the-art end-to-end unsupervised models and most of the multi-stage unsupervised models except the latest method TSD\cite{zhou2023texture}. Meanwhile, our model produces comparable performance with some weakly-supervised and fully-supervised methods. All of the results of the above methods are obtained from public data.

\textbf{Qualitative evaluation.} Shown in Fig. \ref{fig:visual_contrast}, we make the visual comparison with the output of the benchmark methods including weakly-supervised methods MFNet\cite{piao2021mfnet} and SCW\cite{yu2021structure}, multi-stage unsupervised methods SBF \cite{zhang2017supervision}, A2S\cite{zhou2022activation} and TSD\cite{zhou2023texture}, end-to-end methods EDNS\cite{zhang2020learning} and UMNet\cite{wang2022multi}. Our model achieves more accurate results with most of the unsupervised methods and even weakly-supervised models. The shown seven groups of examples illustrate that our module is built with the ability to grab salient objects in different complexity, scales, and depths. Moreover, line 3 shows our model is capable of detecting multi-object saliency, and line 7 shows the effect of our UNSS method. 

\subsection{Pseudo Label Generation}

\begin{table*}
  \caption{Ablation studies on loss functions. The top two results in different methods are marked in red and blue font respectively.}
  \label{tab:loss}
  \begin{tabular}{c|l|ccc|ccc|ccc}
    \toprule
    \multirow{2}[1]{*}{Model} & \multirow{2}[1]{*}{Loss} & \multicolumn{3}{c|}{DUTS-TE} & \multicolumn{3}{c|}{DUT-O} & \multicolumn{3}{c}{ECSSD} \\
    & & $F_\beta \uparrow$ & $E_{\xi} \uparrow$ & $\mathcal{M} \downarrow$ & $F_\beta \uparrow$ & $E_{\xi} \uparrow$ & $\mathcal{M} \downarrow$ & $F_\beta \uparrow$ & $E_{\xi} \uparrow$ & $\mathcal{M} \downarrow$ \\
    \midrule
    A1 & $\mathcal{L}_{pbce}(G)$ + $\mathcal{L}_{lsc}$ & 0.806 & \textcolor{blue}{0.890} & \textcolor{red}{0.052} & 0.743 & 0.845 & 0.074 & \textcolor{blue}{0.902} & \textcolor{red}{0.936} & \textcolor{red}{0.049}\\
    A2 & $\mathcal{L}_{pbce}(G)$ + $\mathcal{L}_{iou}(G)$ & 0.680 & 0.791 & 0.100 & 0.604 & 0.734 & 0.133 & 0.825 & 0.863 & 0.099\\
    A3 & $\mathcal{L}_{pbce}(G)$ + $\mathcal{L}_{iou}(G)$ + $\mathcal{L}_{lsc}$ & \textcolor{red}{0.813} & 0.888 & \textcolor{blue}{0.056} & \textcolor{red}{0.754} & \textcolor{red}{0.854} & \textcolor{blue}{0.070} & \textcolor{blue}{0.902} & 0.920 & 0.059 \\
    \midrule
    B1 & $\mathcal{L}_{pbce}(G_r)$ + $\mathcal{L}_{lsc}$ & 0.730 & 0.826 & 0.082 & 0.686 & 0.789 & 0.112 & 0.845 & 0.912 & 0.065 \\
    B2 & $\mathcal{L}_{pbce}(G_r)$ + $\mathcal{L}_{iou}(G_r)$ & 0.687 & 0.798 & 0.089 & 0.620 & 0.747 & 0.122 & 0.835 & 0.890 & 0.079\\
    B3 & $\mathcal{L}_{pbce}(G_r)$ + $\mathcal{L}_{iou}(G_r)$ + $\mathcal{L}_{lsc}$ & 0.737 & 0.837 & 0.075 & 0.699 & 0.805 & 0.099 & 0.840 & 0.910 & 0.067 \\
    \midrule
    C1 & $\mathcal{L}_{pbce} (G_r)$ + $\mathcal{L}_{iou} (G)$ + $\mathcal{L}_{lsc}$ & 0.733 & 0.829 & 0.076 & 0.681 & 0.785 & 0.106 & 0.843 & 0.911 & 0.067\\
    C2 & $\mathcal{L}_{pbce} (G)$ + $\mathcal{L}_{iou} (G_r)$ + $\mathcal{L}_{lsc}$  & \textcolor{blue}{0.809} & \textcolor{red}{0.891} & \textcolor{red}{0.052} & \textcolor{blue}{0.753} & \textcolor{blue}{0.852} & \textcolor{red}{0.068} & \textcolor{red}{0.903} & \textcolor{blue}{0.935} & \textcolor{blue}{0.050}\\
    \bottomrule
  \end{tabular}
\end{table*}

\begin{table}[!t]
  \centering
\renewcommand\tabcolsep{1pt}
\vspace{-0.1in}
  \caption{Comparison with the equivalent detector.}
  \label{tab:combined}
  \begin{tabular}{c|ccc|ccc}
    \toprule
    \multirow{2}[1]{*}{Methods} & \multicolumn{3}{c|}{ECSSD} & \multicolumn{3}{c}{DUT-O}\\
    & $F_\beta \uparrow$ & $E_{\xi} \uparrow$ & $\mathcal{M} \downarrow$ & $F_\beta \uparrow$ & $E_{\xi} \uparrow$ & $\mathcal{M} \downarrow$\\
    \midrule
    Localizer + CRF + Detector & \textbf{0.909} & 0.922 & 0.056 & \textbf{0.768} & \textbf{0.852} & 0.072\\
    End-to-end detector & 0.903 & \textbf{0.935} & \textbf{0.050} & 0.753 & \textbf{0.852} & \textbf{0.068}\\
    \bottomrule
  \end{tabular}
\end{table}

We employ three-step pseudo labels to learn semantic information at different visual levels. The location pseudos label out the location of the salient objects generally. The detailed pseudos discover the local details and provide more specific semantic guidance. As shown in Tab. \ref{tab:refinement}, the experiments reveal great progress by the refiner and UNSS method in all evaluation measurements. Compared to the location labels, detailed pseudos extract accurate local semantics and precise boundaries. We show the visual changes for the refiner in Fig. \ref{fig:refinement_view}, location labels only focus on the most salient part of the images but detailed labels leverage the detail-boosting activation. And the quality further improved after applying Unsupervised Non-Salient Suppression (UNSS). Moreover, Fig. \ref{fig:unss} shows that the UNSS method not only filters out the non-salient objects in small sizes but also reduces the few-pixel noises that share similar semantics with the salient objects generated by the refiner.

\subsection{Ablation Study}

\textbf{Comparison to an equivalent multi-stage method.} We separate our framework into an almost equivalent multi-stage method to prove that end-to-end multi-task training can not only achieve great performance but also reduce extra overhead. The results shown in Tab. \ref{tab:combined} reveal our combined training framework makes comparable results with an equivalent multi-stage method. In our method, we train the different downstream submodules sharing the same encoder. It is obvious that separating the submodules and training them step by step is quite expensive, so our method avoids the tedious work of training double modules and time-consuming DenseCRF \cite{krahenbuhl2011efficient} especially. Moreover, we surprisingly found that the semantic information was shared between those downstream tasks and even produced fewer prediction errors. 

\textbf{Impact of the hyperparameter $\theta_r$.} We study the impact of hyperparameter $\theta_r$ used in Unsupervised Non-Salient Suppression (UNSS). Hyperparameter $\theta_r$ is employed for determining the size relationship between objects and controlling the probability of attention shift between objects. The smaller $\theta_r$ represents that watcher tends to overlook larger objects in the samples. As shown in Tab. \ref{tab:gamma_r}, we find the hyperparameter $\theta_r = 2.5$ to achieve the best performance. A very large $\theta_r$ makes the UNSS method into a denoise process, which doesn't meet our expectations. Especially, when $\theta_r = +\infty$ represents the situation that detailed labels are used directly without the UNSS method, which is on behalf of a comparison experiment to the method without UNSS. 

\textbf{Effectiveness of loss functions.} Conducting multi-task learning, we propose a complex loss function to supervise the segmentation decoder for several purposes. In that case, we evaluate the best combination of the loss function, the result is shown in Tab. \ref{tab:loss}. For model group A we evaluate the performance only using the location labels for supervision. For model group B only detailed labels are used. Model group C shows the experiments using cross labels for different losses. Comparing experiments in models A2, A3, B2, and B3, we come to the conclusion that the usage of local structure-consistent loss constantly improves the performance on all evaluations. Comparing the same indexed models in different groups shows that for the same loss function, it is better to learn from the most instructive signals in location labels mainly. 
The experiments in model group C show that partial binary cross-entropy loss is better to learn precise regions of the objects and IOU loss fits the learning target of segmenting the complete instances. 
The above observation explains why we set a larger weight for partial binary cross-entropy loss and local structure-consistent loss, but a smaller weight for IOU loss. 
As a result, we discover that partial binary cross-entropy on the localization labels and the IOU loss on the detailed labels is the best choice for ultimate performance. 

\begin{table}
  \centering
\renewcommand\tabcolsep{3pt}
\vspace{-0.1in}
  \caption{Impact of hyperparameter $\theta_r$ in UNSS.}
  \label{tab:gamma_r}
  \begin{tabular}{c|ccc|ccc}
    \toprule
    \multirow{2}[1]{*}{$\theta_r$} & \multicolumn{3}{c|}{ECSSD} & \multicolumn{3}{c}{DUT-O}\\
    & $F_\beta \uparrow$ & $E_{\xi} \uparrow$ & $\mathcal{M} \downarrow$ & $F_\beta \uparrow$ & $E_{\xi} \uparrow$ & $\mathcal{M} \downarrow$\\
    \midrule
    2 & \textbf{0.909} & 0.922 & 0.056 & \textbf{0.768} & \textbf{0.852} & 0.072\\
    2.5 & 0.903 & \textbf{0.935} & \textbf{0.050} & 0.753 & \textbf{0.852} & \textbf{0.068}\\
    3 & 0.907 & 0.929 & 0.053 & 0.759 & 0.850 & 0.075\\
    $+\infty$ & 0.903 & 0.933 & 0.051 & 0.749 & 0.842 & 0.077\\
    \bottomrule
  \end{tabular}
\end{table}

\section{Conclusion}
In this paper, we propose an end-to-end Unsupervised Salient Object Detection (USOD) method by two novel strategies. Taking the idea of top-down context into practice, a multi-level learning model is proposed to learn rich semantic information from global to local. Considering the non-salient emerging issue overlooked by current USOD methods, we propose Unsupervised Non-Salient Suppression (UNSS) to further improve the quality of predictions. As a result, our model produces high-quality saliency maps in an efficient end-to-end pipeline. Extensive experiments demonstrate our method achieves leading performance compared to the current USOD methods. 

\section*{Acknowledgments}

This work was supported by National Natural Science Foundation of China (No.62072112), Scientific and Technological Innovation Action Plan of Shanghai Science and Technology Committee (No.22511102202), National Key R\&D Program of China (2020AAA0108301).

\bibliographystyle{ACM-Reference-Format}
\balance
\bibliography{sample-base}

\end{document}